\documentclass[letterpaper, 10 pt, conference]{ieeeconf}  

\IEEEoverridecommandlockouts                              

\makeatletter
\let\NAT@parse\undefined
\makeatother
\usepackage[colorlinks,
            urlcolor=blue,
            linkcolor=blue,       %
            anchorcolor=blue,  %
            citecolor=green        %
            ]{hyperref}

\usepackage{bm}
\usepackage{graphicx}
\usepackage{amsmath}
\usepackage{amssymb}
\usepackage{amsmath}
\usepackage{multirow}
\usepackage{array}
\usepackage{rotating}
\usepackage{xspace}
\usepackage{subcaption}
\usepackage{color}
\usepackage[dvipsnames]{xcolor}

\newcommand{\name}{MM-Gaussian\xspace}

\title{\LARGE \bf
MM-Gaussian: 3D Gaussian-based Multi-modal Fusion for Localization and Reconstruction in Unbounded Scenes
}

\author{Chenyang Wu, Yifan Duan, Xinran Zhang, Yu Sheng, Jianmin Ji and Yanyong Zhang*
\thanks{* The corresponding author.}
\thanks{School of Computer Science and Technology, University of Science and Technology of China, Hefei, 230026, China
{\tt\small \{cywm39, dyf0202, zxrr, Shengyu724\}@mail.ustc.edu.cn, \{jianmin, yanyongz\}@ustc.edu.cn}.}%
}%

\begin{document}

\maketitle
\thispagestyle{empty}
\pagestyle{empty}

\begin{abstract}
Localization and mapping are critical tasks for various applications such as autonomous vehicles and robotics. The challenges posed by outdoor environments present particular complexities due to their unbounded characteristics. In this work, we present MM-Gaussian, a LiDAR-camera multi-modal fusion system for localization and mapping in unbounded scenes. Our approach is inspired by the recently developed 3D Gaussians, which demonstrate remarkable capabilities in achieving high rendering quality and fast rendering speed. Specifically, our system fully utilizes the geometric structure information provided by solid-state LiDAR to address the problem of inaccurate depth encountered when relying solely on visual solutions in unbounded, outdoor scenarios. Additionally, we utilize 3D Gaussian point clouds, with the assistance of pixel-level gradient descent, to fully exploit the color information in photos, thereby achieving realistic rendering effects. To further bolster the robustness of our system, we designed a relocalization module, which assists in returning to the correct trajectory in the event of a localization failure. Experiments conducted in multiple scenarios demonstrate the effectiveness of our method. 

\end{abstract}

\section{Introduction}

Simultaneous localization and mapping (SLAM) is crucial for the navigation of autonomous systems such as autonomous vehicles and robots~\cite{slamsurvey}, as it can estimate ego-motion and reconstruct the observed scene simultaneously.
In recent years, SLAM techniques have seen considerable progress~\cite{r3live, orb-slam, Pfilter, OCC-VO}, enhancing the precision, speed, and robustness of localization and mapping within SLAM systems significantly.
Despite advancements, traditional SLAM methods are often limited by the map representations, such as point cloud, surfel, and voxel, which can only reconstruct maps at a fixed resolution. This constraint hinders the capture of intricate textures of the scene and prevents SLAM from achieving functionalities like synthesizing new viewpoints.

\begin{figure}[t]
    \vspace{3mm}
    \centering
    \includegraphics[width = \linewidth]{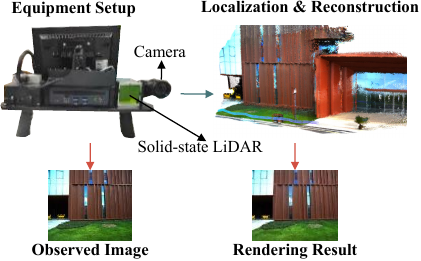}
    \caption{The library reconstructed by \name. We utilize LiDAR and cameras to capture scene data while estimating the sensor pose and reconstructing a 3D Gaussian map in an unbounded scene.}
    \label{fig:head}
    \vspace{-2em}
\end{figure}

Recently, 3D Gaussians~\cite{3dgs} has shown its potential in SLAM technology as a novel map representation. It represents the map as a collection of 3D Gaussian points with attributes such as location, color, and opacity. This representation enables real-time rendering of photorealistic images through a tile-based splatting solution. Simultaneously, by comparing the differences between rendered images and real images, the 3D Gaussians can fully utilize the photometric errors of each pixel and optimize the attributes of Gaussians using gradient descent. Due to these advantages, a series of works applying 3D Gaussians in the SLAM field have emerged~\cite{splatam, gs-slam, GaussianSplattingSlam, Gaussian-slam}, outperforming previous implicit SLAM~\cite{nice-slam, point-slam}. However, these approaches are often based on RGB-D or monocular cameras. 
The absence of depth data in monocular cameras can cause inaccuracies in the placement of 3D Gaussians. Meanwhile, the depth information captured by RGB-D cameras is limited in scope, complicating their use in expansive outdoor scenes.

In this paper, we introduce \name, a novel multi-sensor fusion method, leveraging both LIVOX solid-state LiDAR and a camera. The solid-state LiDAR provides sparse depth information of objects within the camera frustum, and its broad detection range enables our to reconstruct the unbounded outdoor scenes. Specifically, \name efficiently processes LiDAR point clouds and camera images to precisely estimate the trajectory  and  incrementally reconstruct a 3D Gaussian map, which facilitates the rendering of high-quality image. 
Additionally, we observe that degraded scenes in the real world, such as textureless grounds and walls, an result in localization errors, potentially leading to failures in map construction.
Therefore, we design a relocalization module that utilizes the capability of rendering images from Gaussians. This module relocalizes the system once localization fails, guiding it back to the correct trajectory to improve the robustness of \name.

In summary, the main contributions of our work are as follows:
\begin{itemize}
\item We introduce \name, a 3D Gaussians based multi-sensor fusion SLAM method, which utilizes the data from LiDAR and camera. Our system is capable of incrementally constructing a 3D Gaussian map in unbounded, outdoor scenes, and can also render high-quality images in real time.
\item We develop a relocalization module which is designed to correct the system's trajectory in the event of localization failures, thereby enhancing our system's robustness.
\item We record data in a campus scene and test our system. The experimental results show that our method surpasses previous 3D Gaussians SLAM methods in both localization and mapping performance. 
\end{itemize}

\section{Related Works}

\subsection{Neural Radiance Fields based SLAM}
Neural Radiance Fields (NeRF)~\cite{NeRF} is an implicit scene representation method that encodes scenes within the parameters of a neural network. The input to the neural network consists of spatial coordinates and viewing directions, with the output being the volume density and color related to the viewing directions. By sampling along rays in the direction of view, neural radiance fields can utilize a differentiable volume rendering process to render the color of pixels. This allows for the backpropagation of the loss between predicted and actual colors, optimizing the network's parameters through gradient descent. This representation method is notable for its small space occupancy and ability to render high-quality images, attracting many researchers to apply it in the SLAM domain~\cite{imap,nice-slam,point-slam,co-slam}.

The pioneering work in Neural Radiance Fields based SLAM is iMAP~\cite{imap}, which employs a single multi-layer perceptron (MLP) as the map representation. In the tracking stage, it fixes network parameters to optimize pose, while in the mapping stage, it jointly optimizes pose and network parameters, achieving the reconstruction of smaller indoor spaces. Subsequently, NICE-SLAM~\cite{nice-slam} addresses the limitation of a single MLP's representation range by changing the map representation to a collection of hierarchical voxel grids and an MLP decoder. It successfully reconstructs larger spaces by encoding local scene information in the feature vectors corresponding to each voxel. SHINE-Mapping~\cite{Shine-mapping}, using LiDAR point clouds as input, saves space by storing feature vectors in an octree and uses an MLP decoder to predict signed distance values, thus achieving the reconstruction of large outdoor scenes. NeRF-LOAM~\cite{Nerf-loam}, noting the errors in the setting of SDF ground truths in previous methods, divides LiDAR points into ground and non-ground points. This allows for the use of more accurate SDF ground truths to supervise network learning. Although NeRF can render high-quality images and compensate for unseen areas, but the combination of volume rendering and network forward inference limits the system's speed. Moreover, due to neural networks being black-box models, the maps constructed by these methods have poor editability, which limits real-world applications.

\begin{figure*}[thpb]
      \centering
      \includegraphics[width=\textwidth]{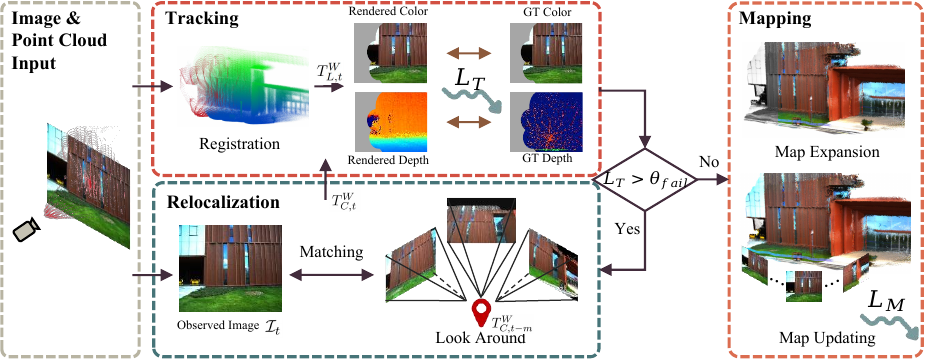}
      \caption{Overview of \name. }
      \label{fig:pipeline}
      \vspace{-2em}
\end{figure*}

\subsection{3D Gaussians based SLAM}

3D Gaussian~\cite{3dgs} is a novel map representation method that, unlike implicit representation methods, discards neural networks and instead uses discrete Gaussian points to represent scenes. Each Gaussian possesses multiple attributes. The images are rendered from the collection of Gaussian points using the splatting. Since there is no network forward inference process, 3D Gaussian can achieve real-time rendering speeds while still maintaining high-quality rendering results. Additionally, 3D Gaussian allows for easier editing, facilitating the application of constructed scenes.

In recent months, several SLAM papers based on 3D Gaussian scene representation have emerged~\cite{splatam,gs-slam,GaussianSplattingSlam,Gaussian-slam}, most of which involve mapping work using RGB-D cameras. SplaTAM~\cite{splatam} introduced a silhouette-guided strategy for adding Gaussian points, which can identify areas in the image where Gaussian points already exist and where they have yet to be added. GS-SLAM~\cite{gs-slam} propose an adaptive 3D Gaussians expansion strategy, along with a coarse-to-fine selection strategy, to efficiently reconstruct scenes and improve camera pose estimation. Gaussian Splatting SLAM~\cite{GaussianSplattingSlam} incorporates isotropic regularization to limit artifacts produced when Gaussian points are generated with insufficient input viewpoints. Although these methods explore the possibility of 3D Gaussian representation used in SLAM, they rely on the depth range of RGB-D cameras and are not suitable for large-scale unbounded scenes. Gaussian Splatting SLAM~\cite{GaussianSplattingSlam} can run in monocular mode, but monocular camera struggle to accurately estimate point depth, leading to inconsistencies between frames when used outdoors. Our method introduces the multi-sensor fusion approach, avoiding these issues by using solid-state LiDAR and camera.

\section{Preliminaries}
Before elaborating on the core components of our \name, it is crucial to outline some foundational algorithms. This section first offers a brief overview of isotropic 3D Gaussian representation, which can reconstruct the structure and texture information of the scene realistically.

3D Gaussians model the scene through a set of 3D Gaussian points, with every Gaussian characterized by multiple learnable attributes. To reduce the learning difficulty, we refer to~\cite{splatam} and introduce isotropic 3D Gaussians, representing each Gaussian with four types of attributes: position $\bm{\mathit{\mu}}_i \in \mathbb{R}^3$, color $\mathbf{c}_i \in \mathbb{R}^3$, opacity $\mathit{o}_i \in [0,1]$, and radius $r_i \in \mathbb{R}^+$. Therefore, each Gaussian is defined as:
\begin{equation}
f(\mathbf{x}) = \mathit{o}_i \exp \left(-\frac{||\mathbf{x}-\bm{\mathit{\mu}}_i||^2}{2r_i^2}\right). 
\end{equation}

To render an RGB image from 3D Gaussians, first for the given camera pose, we sort all Gaussians in front-to-back order and subsequently project them onto the pixel plane. Then, the contribution value $w_i$ of each Gaussian $G_i$ to the color of pixel $\mathbf{p}=(u,v)$ can be calculated as follows:
\begin{equation}
 w_i = f_i(\mathbf{p}) \prod^{i-1}_{j=1} \left(1-f_j(\mathbf{p})\right),   
\end{equation}
where $f_i(\mathbf{p})$ is calculated using (1), but $\bm{\mathit{\mu}}$ and $r$ are the values of the Gaussian projection onto the pixel-space:
\begin{equation}
\bm{\mathit{\mu}}^p = K\frac{E_t\bm{\mathit{\mu}}}{d}, \quad \quad r^p = \frac{fr}{d}, \quad \text{ where }  d = (E_t\bm{\mathit{\mu}})_z.     
\end{equation}
Here, $K$ and $f$ have been given, which represent the camera intrinsic matrix and focal length, respectively. $E_t$ represents the extrinsic matrix of the camera at frame $t$, and $d$ represents the depth of $G_i$ in the camera coordinate system.

With $w_i$, the rendered color $C(\mathbf{p})$ can be written as follows:
\begin{equation}
\label{eq:render_c}
C(\mathbf{p}) = \sum^n_{i=1}\mathbf{c}_i w_i.
\end{equation}
Similarly to~\cite{splatam}, we additionally render the depth map $D(\mathbf{p})$ and the silhouette image $S(\mathbf{p})$ from the 3D Gaussians. The silhouette image represents which pixels in the image contain information from the 3D Gaussian map at the current camera pose. The rendering process for both types of images is:
\begin{equation}
\label{eq:render_d}
D(\mathbf{p}) = \sum^n_{i=1}d_i w_i, 
\end{equation}
\begin{equation}
\label{eq:render_s}
S(\mathbf{p}) = \sum^n_{i=1} w_i. 
\end{equation}

\section{Method}

\subsection{Overview}

Fig. \ref{fig:pipeline} illustrates the overview of \name system, which takes multi-modal data as input, i.e., point clouds from LiDAR and images from the camera, and ultimately outputs a large scale 3D Gaussian map $\mathcal{G}$ that facilitates high-quality image rendering in unbounded scenes. Specifically, the LiDAR and camera both capture point clouds and images at a frequency of 10Hz. At time $t$, point cloud $\mathcal{P}_t^L \in \mathbb{R}^{N\times 3}$ and image $\mathcal{I}_t \in \mathbb{R}^{H\times W\times 3}$ are obtained. Utilizing the extrinsic parameters $C_L^C \in SE(3)$ which are pre-calibrated between the LiDAR and the camera using EdgeCalib~\cite{li2023edgecalib}, the point cloud can be projected onto the image plane, forming a sparse depth image $D_{GT}$ as follow:
\begin{equation}
\label{eq:project}
   d_i = \mathbf{K}C_L^C p_i^L,  
\end{equation}
where the point $p_i$ in $\mathcal{P}_t^L$ is projected onto the camera plane as the pixel coordinate $d_i$. $\mathbf{K}$ is the intrinsic parameters of camera and the subscripts $C$ and $L$ denote the camera and LiDAR coordinate system, respectively.  For notational convenience, we directly use the symbol ``$\cdot$'' to denote the pose transformation of the point cloud, thus excluding the detailed augmentation process.

\name consists of four main components: tracking, relocalization, map expansion, and map updating. During the tracking phase (Sec.~\ref{sec:tracking}), we employ the well-established point cloud registration~\cite{vizzo2023kiss} to obtain initial pose estimates, which are subsequently refined by comparing the camera images with rendered images to enhance the accuracy of the pose estimation. Additionally, to prevent the catastrophic consequences of tracking failures on mapping, a relocalization module (Sec.~\ref{sec:relocalization}) is employed to reset the incorrect position to the correct trajectory, enhancing the robustness of \name in dealing with various scenarios. In the mapping phase (Sec.~\ref{sec:mapping}), point clouds from the LiDAR are processed into 3D Gaussian points and incorporated into the map. Subsequently, a maintained sequence of image keyframes is used to optimize the properties of the 3D Gaussian points in the map, achieving better rendering.

\subsection{Tracking}
\label{sec:tracking}

During the tracking phase, it is essential to determine the sensors' poses to correctly integrate sensor data into the map during the next mapping stage. For constructing a 3D Gaussian map, establishing the camera's pose $T_{C,t}^W$ is crucial, where the superscripts $W$ denote the world coordinate system. However, directly solving for the camera's pose can be challenging or inaccurate due to the lack of 3D information in images. Instead, we first estimate the LiDAR's pose $T_{L,t}^W$ at time $t$, using point cloud registration algorithms~\cite{vizzo2023kiss}. Specifically, for each Gaussian point $G_i$ in the 3D Gaussian map $\mathcal{G}$, we label whether it originates from the LiDAR or is a result of densify, which will be introduced in Sec.~\ref{sec:mapping}.
We retain only the position attributes of LiDAR-originated points, treating them as a normal point cloud, which are then registered with the LiDAR points $P_t^L$ at time $t$ to obtain the LiDAR's pose $T_{L,t}^W$.

Subsequently, the camera's pose is derived as $T_{C,t}^W = T_{L,t}^W \cdot C_C^L$. Utilizing this pose, we render the RGB, depth and silhouette images using Eq.~\eqref{eq:render_c}, Eq.~\eqref{eq:render_d} and Eq.~\eqref{eq:render_s} from the 3D Gaussian map $\mathcal{G}$.
Next, we further optimize the camera's pose by comparing the differences between the rendered RGB and depth images with the captured RGB image and depth image generated by the projection.
Since the images contain parts that have not yet been reconstructed, in order to avoid accumulating loss in this part and thus affecting pose optimization, we follow the approach described in \cite{splatam} and establish a threshold $\theta_s$. We consider regions within the silhouette image that exceed this threshold to be already reconstructed, and consequently, we only calculate loss within these areas. Our loss function is composed of color loss and depth loss, both of which utilize the L1 norm. By applying a weighting factor $\lambda_c$, we combine color loss and depth loss, and employ the Adam optimizer~\cite{adamOptimizer} for gradient descent to determine the estimated pose $\hat{T}_{C,t}^W$ of the current frame. The loss is as follow:
\begin{equation}
\label{eq:loss}
L_T = \mathbf{\sum_p}\left(S(\mathbf{p}) > \theta_s\right)\left(L_1\left(D(\mathbf{p})\right) + \lambda_c L_1\left(C(\mathbf{p})\right)\right). 
\end{equation}

\begin{figure}[t]
      \centering
      \includegraphics[width=0.45\textwidth]{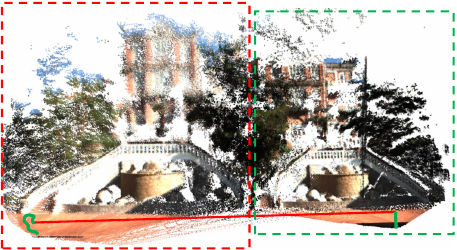}
      \caption{The failure in map reconstruction due to the failure in tracking. When tracking is successful, mapping proceeds smoothly, as illustrated by the map and green trajectory in the  green box. However, a segment of data recorded facing the ground leads to trajectory drift indicated by the red trajectory. Subsequently, tracking stabilizes, but by this time, the map reconstructed within the red box diverged from the correct map location shown in the green box.}
      \label{fig:groundAndWall}
      \vspace{-2em}
\end{figure}

\subsection{Relocalization}
\label{sec:relocalization}
\begin{figure}[t]
      \centering
      \includegraphics[width=0.45\textwidth]{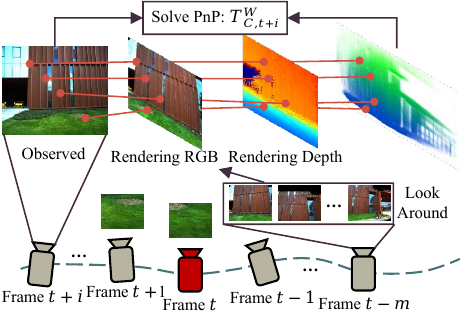}
      \caption{Tracking failed at the $t$ th frame. We use the $t-m$ th frame as a recovery point to perform a look-around operation. By solving the PnP problem, the pose of $t+i$ th frame is estimated successfully.}
      \label{fig:relocalPipeline}
      \vspace{-2em}
\end{figure}

Pose estimation is not always successful, particularly when dealing with degenerate scenes such as walls and floors.  Tracking failures can significantly impact the reconstruction of the 3D Gaussian map, as illustrated in Fig.~\ref{fig:groundAndWall}. To address this issue, we first introduce a tracking failure detection module and then reset the incorrect pose back to the correct trajectory  through a relocalization module.

In tracking failure detection, we assess whether the loss calculated using Eq.~\eqref{eq:loss} for each frame exceeds the threshold $\theta_{fail}$. If the loss surpasses $\theta_{fail}$, the \name system will be set to a tracking failure state. In this state, the tracking module is considered unable to continue outputting the correct pose and ceases accepting new data. Consequently, map expansion and map updating processes are also halted. Simultaneously, the relocalization module is activated.

As illustrated in Fig.~\ref{fig:relocalPipeline}. Specifically, upon failure at frame $t$, we retrieve the camera's pose $T_{C,t-m}^W$ from $m$ frames prior, which is considered a correct pose. We keep the translational part of $T_{C,t-m}^W$ unchanged and perform a ``look-around'' operation on the rotation, i.e., uniformly sampling $n$ rotations to form $n$ new poses. The corresponding RGB, depth and silhouette images of the $n$ poses are also rendered. Then, for each frame $\mathcal{I}_{t+i}$ captured by the camera after the tracking failure, we utilize SuperPoint~\cite{detone2018superpoint} for feature extraction  and LightGlue~\cite{lindenberger2023lightglue} for feature matching between the current frame and the $n$ rendered RGB images. We select one of the $n$ images with the highest number of matching points while exceeding the threshold $\theta_{feature}$ as the candidate. Using the pose of the candidate, we project the rendered depth map back into 3D space via the inverse of Eq.~\eqref{eq:project} and then calculate the current frame's pose $T_{C,t+i}^W$ using Perspective-n-Point (PnP) based on the correspondence of features. With the outcome, we render the corresponding RGB, depth and silhouette image again and assess its loss by Eq.~\eqref{eq:loss}. If the loss is below the threshold $\theta_{fail}$, relocalization is considered successful. Tracking, map expansion, and map updating module will be resumed. Frames between the failed t-th frame and the successfully relocated t+i-th frame will be discarded to avoid impacting the 3D Gaussian map.

\begin{table*}[h]
\caption{Quantitative Evaluation of Tracking. (ATE RMSE $\downarrow$ [m])}
\label{table:tracking}
\begin{center}
\begin{tabular}{l|cccccccccc}
\hline
Methods    & 00      & 01       & 02       & 03      & 04     & 05     & 06  & 07 & 08 & Mean \\
\hline
SplaTAM~\cite{splatam}   & 18.31   & 158.9    & 156.1    & 14.73   & 7.312  & 3.649  & 7.086 & 1.844 & 123.2 & 54.57 \\
\hline
MonoGS~\cite{GaussianSplattingSlam}    & 21.02   & 16.59    & 13.78    & 18.94   & 6.165  & 19.05  & 9.191 & 3.174 & 15.03 & 13.66 \\
\hline
NeRF-LOAM~\cite{Nerf-loam} & 1.782   & 3.283    & 0.286    & 0.788   & 1.257  & 0.718  & 1.979  & 0.555 & 0.377 & 1.225 \\
\hline
ours      & 0.088   & 0.243    & 0.071    & 0.217   & 0.272  & 0.241  & 0.326  & 0.488  & 0.299 & 0.249 \\
\hline
\end{tabular}
\end{center}
\vspace{-1em}
\end{table*}

\begin{table*}[h]
\caption{Quantitative Evaluation of Mapping.}
\label{table:mapping}
\begin{center}
\begin{tabular}{lc|c|cccccccccc}
\hline
Methods                      &Pose     & Metrics & 00    & 01    & 02    & 03    & 04    & 05    & 06    & 07    & 08    & Mean  \\
\hline
\multirow{3}{*}{SplaTAM~\cite{splatam}}     &          & PSNR$\uparrow$    & 21.28 & 19.62 & 16.72 & 20.18 & 17.85 & 16.20 & 19.28 & 20.51 & 15.82 & 18.61 \\
                             &GT pose   & SSIM$\uparrow$    & 0.747 & 0.685 & 0.623 & 0.737 & 0.611 & 0.590 & 0.725 & 0.726 & 0.624 & 0.674 \\
                             &          & LPIPS$\downarrow$   & 0.495 & 0.484 & 0.561 & 0.454 & 0.542 & 0.609 & 0.615 & 0.551 & 0.523 & 0.537 \\
\hline
\multirow{3}{*}{MonoGS~\cite{GaussianSplattingSlam}}      &          & PSNR$\uparrow$    & 12.34 & 8.435 & 8.615 & 9.413 & 8.149 & 9.109 & 11.00 & 13.54 & 9.247 & 9.983 \\
                             &GT pose   & SSIM$\uparrow$    & 0.078 & 0.038 & 0.035 & 0.036 & 0.029 & 0.039 & 0.067 & 0.104 & 0.034 & 0.051 \\
                             &          & LPIPS$\downarrow$   & 0.935 & 0.911 & 1.051 & 0.938 & 0.957 & 0.980 & 1.037 & 0.974 & 1.025 & 0.979 \\
\hline
\multirow{3}{*}{3D Gaussian Splatting~\cite{3dgs}} & & PSNR$\uparrow$    & 15.30 & 13.75 & 12.72 & 14.16 & 14.66 & 11.98 & 15.08 & 17.15 & 11.33 & 14.01 \\
                             &GT pose   & SSIM$\uparrow$    & 0.668 & 0.631 & 0.625 & 0.705 & 0.606 & 0.575 & 0.718 & 0.722 & 0.588 & 0.649 \\
                             &          & LPIPS$\downarrow$   & 0.666 & 0.707 & 0.792 & 0.705 & 0.743 & 0.776 & 0.678 & 0.679 & 0.802 & 0.728 \\
\hline
\multirow{3}{*}{ours}        &          & PSNR$\uparrow$    & 24.54 & 21.76 & 21.60 & 22.16 & 20.48 & 18.81 & 21.81 & 21.94 & 17.51 & 21.18 \\
                             &Odometry  & SSIM$\uparrow$    & 0.875 & 0.802 & 0.831 & 0.823 & 0.737 & 0.758 & 0.898 & 0.863 & 0.805 & 0.821 \\
                             &          & LPIPS$\downarrow$   & 0.138 & 0.223 & 0.172 & 0.206 & 0.304 & 0.231 & 0.104 & 0.143 & 0.237 & 0.195 \\
\hline
\end{tabular}
\end{center}
\vspace{-2em}
\end{table*}

\subsection{Mapping}
\label{sec:mapping}
\subsubsection{Map Expansion}
\label{sec:map_expansion}

After the tracking stage, we obtain the estimated pose of the current frame. Based on this pose, we transform the LiDAR point cloud of the current frame into 3D Gaussians and add them to the map for expansion.
Specifically, for the incoming $N$ points in $\mathcal{P}_t^L$, we convert them into $N$ Gaussian points and add them to the map using the refined LiDAR pose:
\begin{equation}
    \hat{T}_{L,t}^W = \hat{T}_{C,t}^W \cdot C_L^C,~  \mathcal{P}_t^W= \hat{T}_{L,t}^W \cdot \mathcal{P}_t^L.
\end{equation}

The position $\bm{\mathit{\mu}}$ of the Gaussian point $G$ added to the 3D Gaussian map $\mathcal{G}$ is determined by the corresponding position of $p$ in $\mathcal{P}_t^W$.
By projecting the point cloud onto the pixel plane, we can use the colors of the projected pixels as the initial colors for the Gaussians. For the new Gaussian point $G_i \in \mathcal{G}$, the radius of it is initialized as 
$r_{G_i} = \frac{D_{G_i}}{f}$,
where $f=\frac{f_x+f_y}{2}$, $f_x$ and $f_y$ are the the focal length of the camera in the $x$ and $y$ directions, respectively. $D_{G_i}$ represents the depth value  in the camera coordinate system, which is the distance from the point to the camera. The opacity of each Gaussian is initialized to a constant value of 0.5.

\subsubsection{Map Updating}
\label{sec:map_updating}
We maintain a continuously growing keyframe sequence $Q_{key}$, adding a new frame to the sequence after a certain number of input frames. Before the mapping stage, we select $k-2$ frames from $Q_{key}$ to optimize together with the current frame and the latest keyframe in this mapping stage. To select the keyframes most relevant to the current frame, we transform the point cloud of the current frame into the world coordinate system, and then project it onto each keyframe. The selection is based on the number of points projected onto the pixel plane.

During the mapping stage, our goal is to update the attributes of the 3D Gaussians, without optimizing the camera pose. Therefore, we set a fixed number of iterations, each time randomly selecting one frame from the previously chosen $k$ frames. Based on the estimated camera pose from selected frame, we render an RGB image and then compute the loss function according to the input images. 

Since we have already transformed the input LiDAR point cloud into Gaussians and added them to the map, we no longer include depth loss in the loss function of the mapping stage. Instead, we add SSIM~\cite{SSIM} (Structural Similarity Index Measure) loss and continue to use the Adam optimizer~\cite{adamOptimizer}. The final loss function is as follows: 
\begin{equation}
\label{eq:lossMapping}
L_M = \left(1-\lambda_{S}\right)\cdot|\hat{\mathcal{I}}-\mathcal{I}|+\lambda_{S}\left(1-\text{SSIM}(\hat{\mathcal{I}}, \mathcal{I})\right), 
\end{equation}
where $\hat{\mathcal{I}}$ is the rendered image, $\mathcal{I}$ is the original image.   

In the optimization process, some useless Gaussians may become transparent, or too large. Therefore,  according to the 3D gaussian splatting~\cite{3dgs}, we add a pruning step for Gaussians at the end of the mapping stage, removing these useless Gaussians. In addition to this, to more finely represent the details of object surfaces, we incorporate a densification process, which involves duplicating Gaussians based on gradients to generate new Gaussians as done in~\cite{3dgs}.

\begin{figure*}[t]
      \centering
      \includegraphics[width=0.85\textwidth]{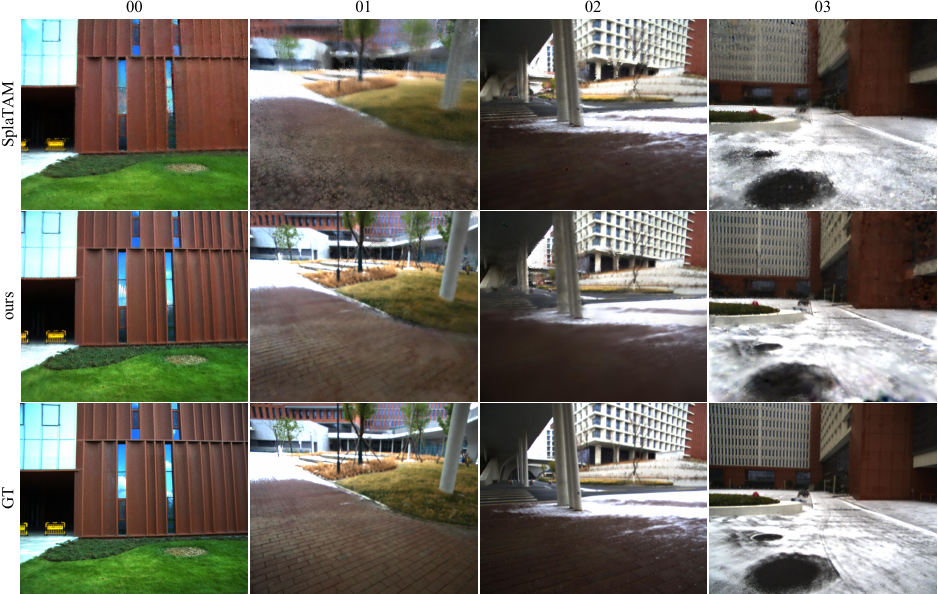}
      \caption{Rendering results of SplaTAM~\cite{splatam} and \name.}
      \label{fig:qualitativeResult}
      \vspace{-2em}
\end{figure*}

\section{Experiment}

\subsection{Experimental Setup}

We assemble a data collection device using Livox AVIA LiDAR\footnote{https://www.livoxtech.com/avia} and MV-CS050-10UC camera\footnote{https://www.hikrobotics.com/en/machinevision/productdetail?id=6338}, and  collect nine data sets in the USTC campus. All experiments are conducted using the parameters setting in the Tab.~\ref{table:ParameterSetting}.
For the pose ground truth, we initially use R3LIVE~\cite{r3live}, the state-of-the-art multi-model SLAM system, to obtain the preliminary pose. 
Subsequently, we employ HBA~\cite{hba}, an offline tool for optimizing map quality, to further enhance the accuracy of the trajectory.
Additionally, to evaluate our proposed relocalization module, we also use data from the R3LIVE dataset, which includes many challenging scenarios.

\begin{table}[t]
\caption{Parameter Settings.}
\label{table:ParameterSetting}
\begin{center}
\begin{tabular}{cccccccc}
\hline
$\theta_s$ & $\lambda_c$ & $\theta_{fail}$ & $m$  & $n$ & $\theta_{feature}$ & $k$  & $\lambda_{S}$ \\
\hline
0.99       & 0.5         & 1e+5            & 30 & 8 & 10                 & 20 & 0.2   \\
\hline
\end{tabular}
\end{center}
\vspace{-2em}
\end{table}

We select several methods for comparison with ours. First, we consider 3D Gaussian-based SLAM methods such as SplaTAM~\cite{splatam} and Gaussian Splatting SLAM~\cite{GaussianSplattingSlam} (abbreviated as MonoGS). Secondly, when evaluating the tracking component, we also choose the NeRF-LOAM~\cite{Nerf-loam}, which uses LiDAR point clouds as input and is based on Neural Radiance Fields. For the evaluation of the mapping component, we additionally select 3D Gaussian Splatting~\cite{3dgs}.

For evaluation metrics of tracking, we present the root-mean-square error (RMSE) of the absolute trajectory error (ATE). For evaluation metrics of mapping, we report peak signal-to-noise ratio (PSNR), SSIM~\cite{SSIM} and LPIPS~\cite{LPIPS}.

\subsection{Evaluation of Tracking}

Tab.~\ref{table:tracking} shows the results of our comparison of tracking module. SplaTAM~\cite{splatam} and MonoGS~\cite{GaussianSplattingSlam} use RGB-D or monocular camera as input. Therefore, we transform the LiDAR point cloud to the camera coordinate system first, then project it onto the pixel plane to obtain a depth image, which serves as the input for these two methods. The point cloud obtained by AVIA LiDAR scanning is concentrated in front of the sensor, with the majority of points located within the camera frustum and can be projected to generate a depth image. Hence, in terms of the density of depth information provided as input, we are the same as SplaTAM~\cite{splatam} and MonoGS~\cite{GaussianSplattingSlam}. However, as seen in the Tab.~\ref{table:tracking}, we achieve significantly better tracking results than these two methods. The first reason is that we do not use a constant velocity motion model to provide the initial pose value, but instead solve ICP for the LiDAR point cloud to obtain a better initial value. The second reason is that when there are large and rapid turns or movements in the input data, tracking results are prone to errors, and our proposed relocalization module can capture such cases of localization failure, promptly bringing the trajectory back to the correct path, thereby improving tracking accuracy. NeRF-LOAM~\cite{Nerf-loam} utilizes LiDAR point clouds for reconstructing outdoor large scenes, hence yielding better tracking results than SplaTAM~\cite{splatam} and MonoGS~\cite{GaussianSplattingSlam}. However, under good initial conditions, 3D Gaussians can render images and further optimize the pose using information from each pixel, allowing us to achieve better results than NeRF-LOAM~\cite{Nerf-loam}.

\subsection{Evaluation of Mapping}
Tab.~\ref{table:mapping} shows the results of our comparison of mapping module. SplaTAM~\cite{splatam} and MonoGS~\cite{GaussianSplattingSlam} use the same depth images as input as described earlier. Considering the significant errors in their tracking results, we utilize ground truth poses as their odometry. 3D Gaussian Splatting~\cite{3dgs} does not perform pose estimation, thus also employing ground truth poses. Additionally, our method still utilizes our own odometry. Furthermore, to expedite the convergence of 3D Gaussian Splatting~\cite{3dgs}, we concatenate every LiDAR point cloud frame using ground truth poses to form a complete point cloud map. This map is then used as the initial value for Gaussian positions in this method. From the table, it can be seen that our method achieves the best results across all sequences. We convert all LiDAR point clouds obtained from each frame into 3D Gaussians, thereby fully utilizing the non-repetitive scanning mode of AVIA LiDAR. As the system scans a particular area over time, the coverage area ratio increases, enabling Gaussians to represent the surface of objects in finer detail.

Fig. \ref{fig:qualitativeResult} shows the qualitative results of the four data sequences in our dataset. For comparison, it also presents SplaTAM~\cite{splatam}, which has the highest rendering quality amongst the other three methods. It can be observed that the images rendered by our method contain less blur and present a better representation of the detailed parts of the object's surface. Moreover, areas such as the edges of tree trunks and pillars are also depicted with clearer boundaries.

\subsection{Ablation Study for Relocalization}
The proposed relocalization module in our study is designed to detect failures in tracking and attempt to recover. To evaluate the effectiveness of the relocalization module, we used two sequences of data from the R$^3$LIVE dataset~\cite{r3live} (marked as deg00 and deg01 in the Tab.~\ref{table:ablation}). These two data sequences include sensor degeneration due to LiDAR's proximity to the ground and to the wall, respectively. In addition, we record data facing the wall in an indoor corridor (marked as 09 in the Tab.~\ref{table:ablation}), which can influence the LiDAR point cloud through turning close to the wall surface. These three data sequences will cause ICP to be unable to calculate a more accurate pose, or even completely deviate from the original path. In the Tab.~\ref{table:ablation}, we list the ATE RMSE obtained without relocalization and with relocalization (calculated by comparing the pose of the frames after successful relocalization with the Ground Truth pose). The results show that our relocalization module has successfully restored the pose to the correct path.

\begin{table}[t]
\caption{Ablation Study for Relocalization.}
\label{table:ablation}
\begin{center}
\begin{tabular}{c|ccc}
\hline
ATE RMSE$\downarrow$[m]  & deg00 & deg01 & 09    \\
\hline
w/o Relocalization & 13.08 & 12.59 & 2.655 \\
\hline
w/ Relocalization  & 0.204 & 0.513 & 0.274 \\
\hline

\end{tabular}
\end{center}
\vspace{-2em}
\end{table}

\section{Conclusion}

In this paper, we propose a 3D Gaussians-based multi-sensor fusion SLAM method, \name. Existing 3D Gaussians SLAM methods often utilize RGB-D or monocular cameras, which limits their applicability. Therefore, we employ Livox AVIA LiDAR and a camera as sensor inputs, achieving online localization and construction of a 3D Gaussian map. Considering the complexity of real-world environments, we introduce a relocalization module that leverages the rendering capabilities of 3D Gaussians. It attempts to restore the pose to the correct trajectory in case of localization failures, thereby enhancing the robustness of our system. This is one of the earliest studies that explores 3D Gaussians SLAM in outdoor unbounded scenes. Moving forward, we will further improve the accuracy of localization and mapping, as well as increase the system's speed.

\addtolength{\textheight}{-12cm}   









\bibliographystyle{IEEEtran}
\bibliography{IEEEabrv, references}

\end{document}